\pdfoutput=1

\documentclass[11pt]{article}

\usepackage[preprint]{acl}
\usepackage{float}

\usepackage{times}
\usepackage{latexsym}
\usepackage{booktabs}
\usepackage{amsmath}
\usepackage{multirow}
\usepackage{amssymb}
\usepackage[T1]{fontenc}

\usepackage[utf8]{inputenc}

\usepackage{microtype}

\usepackage{inconsolata}

\usepackage{graphicx}

\title{GFRIEND: Generative Few-shot Reward Inference through EfficieNt DPO}

\author{
    Yiyang Zhao$^{1}$\and
    Huiyu Bai$^{1}$\and
    Xuejiao Zhao$^{1,2}$\thanks{Corresponding Author}\\
    \textsuperscript{1}College of Computing and Data Science,Nanyang Technological University (NTU), Singapore\\
    \textsuperscript{2}Joint NTU-UBC Research Centre of Excellence in Active Living for the Elderly(LILY),NTU,Singapore\\
    \{YIYANG004,huiyu001\}@e.ntu.edu.sg,\,xjzhao@ntu.edu.sg
}

\begin{document}
\maketitle
\begin{abstract}
The ability to train high-performing reward models with few-shot data is critical for enhancing the efficiency and scalability of Reinforcement Learning from Human Feedback (RLHF). We propose a data augmentation and expansion framework that enables generative reward models trained on small datasets to achieve comparable performance to those trained on large-scale datasets. Traditional methods to train a generative reward model, such as Direct Preference Optimization (DPO), are constrained by inefficiencies in sample pairing and limited data diversity. This work introduces preference refinement, which employs Chain-of-Thought (CoT) sampling to uncover diverse and high-quality preference relationships. It also incorporates a perplexity-based scoring mechanism to assign nuanced preference levels and utilizes Multi-level Direct Preference Optimization (M-DPO) to enable the model to capture finer-grained preference differences between samples. Experimental results demonstrate that the proposed method significantly enhances data efficiency and model performance, enabling reward models trained in a few-shot setting to achieve results on par with those trained on large-scale datasets. This study underscores the potential of data-efficient strategies in advancing reward model optimization, offering a robust solution for low-resource RLHF applications. See our code: https://github.com/SNOWTEAM2023/GFRIEND
\end{abstract}

\section{Introduction}

Large language models (LLMs) have achieved remarkable successes in a variety of natural language processing (NLP) tasks \citep{ouyang2022training, rafailov2023direct, schulman2017proximal, bai2022a}. However, aligning these models with human values and preferences remains a fundamental challenge. RLHF has emerged as a key approach to optimizing model behavior, and the core of this method is to train a reward model using human-labeled preference data \citep{christiano2017deep, rafailov2023direct, askell2021general}. This reward model is capable of predicting human preferences based on the training data, thereby guiding the optimization of policies to align LLMs.  Existing frameworks typically rely on pairwise preference data to train reward models. However, these methods suffer from inefficiencies in sample pairing and limited diversity in preference data, leading to suboptimal generalization. Traditional reward modeling approaches, such as the Bradley-Terry (BT) model \citep{bradley1952rank}, have been widely used for preference estimation. However, they struggle to capture complex and intransitive human preferences, which can result in biases such as verbosity preference, where models favor longer responses irrespective of quality \citep{ouyang2022training, lambert2024rewardbench}. Furthermore, common RLHF strategies, including Proximal Policy Optimization (PPO) \citep{schulman2017proximal} and reward-ranked fine-tuning (RAFT) \citep{dong2023raft}, heavily depend on the robustness of reward models, making them a crucial component in the alignment pipeline. Recent studies have proposed alternative strategies, such as multi-objective reward modeling and iterative preference learning, to improve the interpretability and adaptability of reward models \citep{rafailov2023direct, xu2023online}. Despite these advancements, the utilization of these techniques in low-data environments remains an open challenge. In specialized domains such as healthcare, law, and finance, the collection of large-scale preference data is often significantly constrained. Recent works have explored data augmentation \cite{hosseini2024vstar,luong2024reft} and meta-learning\cite{zhang2024prototypical} to enhance models' generalization ability in low-data settings. 

\begin{figure}
\centering
\includegraphics[width=0.48\textwidth]{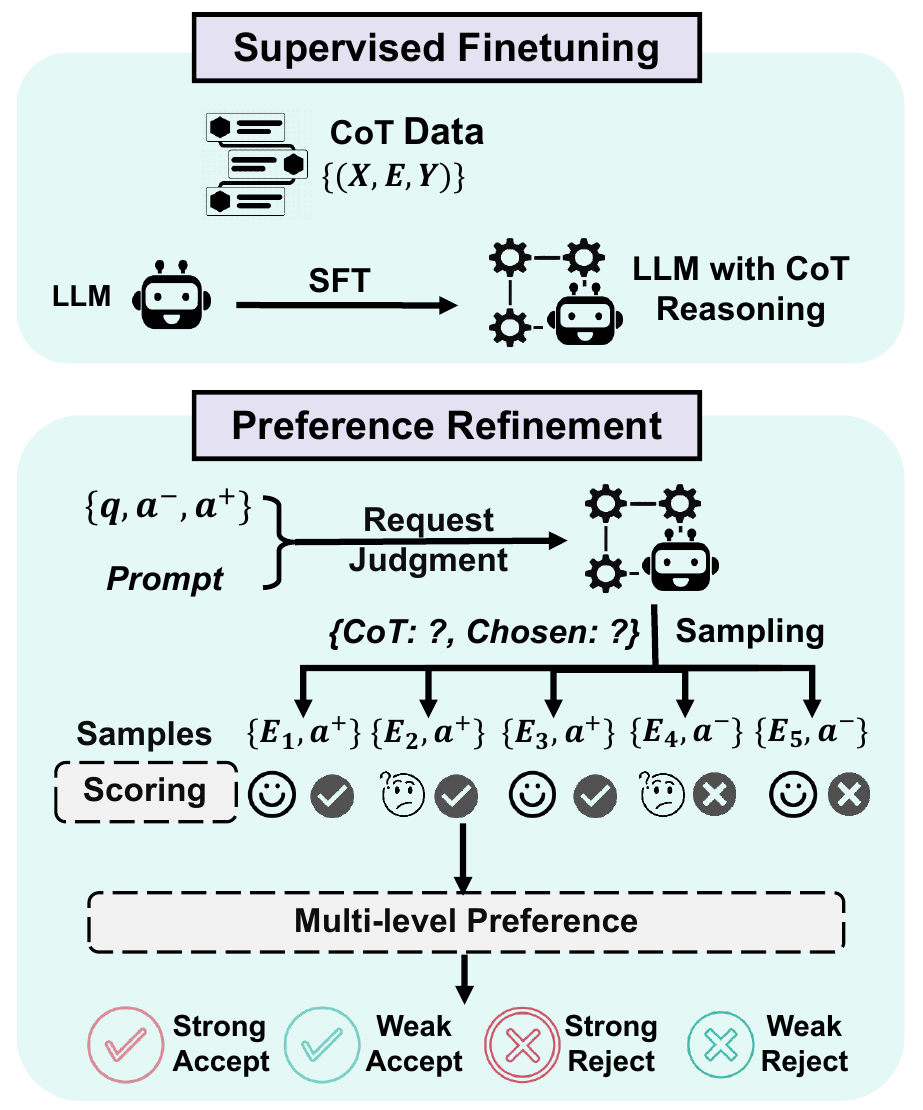}
\caption{The steps for enhancing LLM with CoT Reasoning and CoT sampling with a preference dataset that includes preference labels for a pair of answers to a question.} 
\label{fig1}
\end{figure}

To address these challenges, we introduce a data augmentation and generative reward modeling framework for RLHF that efficiently leverages limited preference data. Our approach integrates preference refinement module to produce diverse, high-quality preference data (see Figure \ref{fig1}), mitigating data sparsity \citep{wei2022chain, hosseini2024vstar}. We employ a multi-level preference modeling strategy rather than simple binary comparisons, using a perplexity-based scoring mechanism to quantify preference degrees and enable finer-grained reward modeling. Finally, we modify the Direct Preference Optimization (DPO) loss by weighting sample pairs based on preference disparity, ensuring more representative data is emphasized during reward model training.

We evaluate this framework on multiple benchmarks, comparing it against conventional reward modeling approaches. Our experiments confirm that data augmentation improves preference modeling accuracy and that multi-level preference scoring is effective. Ablation studies reveal the contributions of each component. The results show significant performance gains in low-resource settings, comparable to models trained on larger datasets, underscoring the potential of data-efficient strategies for advancing RLHF optimization. 

Our contributions can be summarized as follows:
\begin{itemize}
    \item \textbf{Efficient Data Augmentation:} We introduce a Chain-of-Thought (CoT) sampling mechanism to generate diverse and high-quality preference data, improving the robustness of reward models under low-resource conditions \citep{wei2022chain, hosseini2024vstar}.
    \item \textbf{Multi-Level Preference Modeling:} We propose a perplexity-based scoring method to assign nuanced preference levels, enabling more fine-grained reward model training compared to traditional binary pairwise approaches.
    \item \textbf{Optimized Preference Learning:} We enhance the DPO loss function by incorporating weighted sample pairs based on preference disparities, leading to improved generalization and stability in low-data settings.
    \item \textbf{Empirical Validation:} We conduct extensive experiments demonstrating that our method significantly enhances data efficiency and achieves performance comparable to models trained on large-scale datasets, underscoring its potential for scalable and resource-efficient RLHF applications.
\end{itemize}

\section{Related Works}

Reward modeling is crucial for RLHF, as it aligns LLMs with human preferences by predicting preference scores for sample pairs \citep{christiano2017deep, rafailov2023direct, askell2021general}. A commonly used approach is the Bradley-Terry (BT) model \citep{bradley1952rank}, which infers the likelihood of one sample being preferred over another. Despite widespread adoption, its reliance on large labeled datasets restricts applicability in data-scarce fields like healthcare and law \citep{ouyang2022training, lambert2024rewardbench}.

To mitigate data scarcity, researchers have explored data augmentation (e.g., chain-of-thought reasoning, probabilistic sampling) \citep{wei2022chain, hosseini2024vstar, luong2024reft}, which generates diverse rationales for preference judgments \citep{wei2022chain,kojima2022large,zhang2022automatic}. Multi-level preference modeling can also capture finer distinctions between samples \citep{ye2024beyond, kim2024prometheus}. In optimizing reward models with limited data, a weighted mechanism is often employed to prioritize high-quality samples \citep{zhang2024prototypical, rafailov2023direct}. For instance, the Prototypical Reward Network (Proto-RM) introduces a prototypical network for data-efficient RLHF \citep{zhang2024prototypical}, while perplexity-based loss weighting effectively improves performance \citep{rafailov2023direct}. Active Preference Optimization (APO) \citep{das2024active} further enhances model alignment under small sample budgets, highlighting the need for scalable, data-efficient solutions \citep{wang2024reinforcement, dong2023survey}.

Generative data and generative reward modeling have also gained traction. Generative feedback can be applied to construct or expand preference datasets \citep{ye2024beyond} and support online algorithms such as iterative-DPO or online-DPO \citep{dong2024rlhf,xu2023online,xiao2024comprehensive}. Concurrently, specialized evaluators are being developed: for example, prompting GPT-4 to produce instruction-tuning data helps train generative judges \citep{li2023instruction, kim2024prometheus}, and minimal “Yes/No” supervision with external rationales further refines these models \citep{zhang2024fine}. GenRM and CoT-GenRM use next-token prediction to compute preference probabilities without altering model architectures \citep{mahan2024generative}, while integrating both positive and negative preference data can enhance evaluation \citep{wang2024direct}. Lastly, CLoud \citep{ankner2024critiqueoutloud} first elicits a comment on the response, then outputs a reward by incorporating the prompt, response, and generated comment.

\section{Method}

\subsection{Preliminary}

Given a query $q$ and a pair of responses $a_1$ and $a_2$, the objective is to determine the preferred response between $a_1$ and $a_2$. To achieve this, the model is trained on a preference dataset $D = \{(q, a^-, a^+)_i\}_{i=1}^{N}$, where $a^+$ represents the preferred response relative to $a^-$. The model's performance is tested by checking its accuracy on a separate and non-overlapping dataset.

\noindent
\textbf{Reward Modeling in RLHF}

A widely adopted method of preference determination is Reward Modeling (RM), which plays a crucial role in RLHF \cite{christiano2017deep, ouyang2022training}. RM is used to infernumerical scores that reflect the relative quality of different responses based on human-labeled preference data. It is also called a scalar model, which is typically initialized by concatenating a pre-trained LLM with a randomly initialized MLP head. This approach enables a reward model to provide scalar feedback, guiding policy updates in reinforcement learning.

In a standard setting, the RM assigns a score $r(q, a)$ for each response $a \in \{a_1, a_2\}$ and infers preference by comparing $r(q, a_1)$ and $r(q, a_2)$. The training objective of RM is often based on the Bradley-Terry (BT) model \cite{bradley1952rank}, which maximizes the probability that the preferred response $a^+$ is ranked higher than $a^-$:

\begin{align}
P (a^+ \succ a^-|q) = \frac{\exp(r(q, a^+))}{\exp(r(q, a^+)) + \exp(r(q, a^-))}  \nonumber \\
=\sigma(r(q, a^+) - r(q, a^-)),
\end{align}
where $\sigma(\cdot)$ denotes the sigmoid function.

While effective, traditional reward modeling often require large-scale human-annotated data, making them less suitable for low-resource domains such as healthcare, law, and finance, where data collection is expensive and time-consuming \cite{lambert2024rewardbench, wang2024reinforcement}. Moreover, human preferences are not always transitive or deterministic, introducing additional challenges in preference modeling.

\noindent
\textbf{LLM-Based Generative Preference Judgment}

Recent research has demonstrated that large language models (LLMs) themselves possess strong capabilities for preference evaluation, reducing reliance on explicit reward models \cite{kojima2022large, wei2022chain}. In this alternative approach, the LLM directly generates a preference judgment given a structured prompt:

\begin{equation}
p = T (q, a_1, a_2),
\end{equation}
where $T (\cdot)$ is a predefined prompt template that integrates the query and candidate responses into a standardized format. The LLM then processes $p$ and generates a judgment $j$:

\begin{equation}
j = \pi(p),
\end{equation}
where $\pi(\cdot)$ represents the LLM’s policy function that outputs a natural language judgment, potentially including explanations supporting the decision. This method leverages the \textbf{in-context learning} ability of LLMs, allowing them to make informed judgments based on implicit knowledge acquired during pretraining~\cite{zhao2021brain,zhao2021explainable}.

\subsection{GFRIEND Framework}

\noindent \textbf{Enhance LLM with CoT Reasoning}

The base model is fine-tuned for epochs on a dataset comprising tuples of “$(question, CoT)$,” denoted as $(x,e)$. Formally, the CoTs' generation can be decomposed into a sequence of next-token predictions. The final prediction is marked with the $<eos>$ token, indicating the end of the generation process. The CoT e can be represented as:

\begin{equation}
e= \begin{bmatrix} a_1,a_2,...,a_{L-1},a_L\mathrm{=<eos>} \end{bmatrix},    
\end{equation}
where $L$ denotes the maximum sequence length. At each time step $t$, the action $a_t$ is sampled from the policy $\pi_\theta(\cdot|s_t)$, where $a_t$ can be any token in the vocabulary, and the state $s_t$ includes all tokens in the question and all tokens generated so far. After each action, the resulting state $s_{t+1}$ is the concatenation of the current state $s_t$ and the action $a_t$:
\begin{equation}
s_{t+1}= \begin{cases} \boldsymbol{x}, & t=0 \\ [s_t,a_t], & 1\le t\le L \end{cases}    
\end{equation}

When the action generated is $<eos>$, the resulting state $s_{L+1}$ is the terminal state, and the generation process terminates. With this formulation, the loss function can be written as:

\begin{equation}
\mathcal{L}_{SFT}(\boldsymbol{\theta})=-{E}_{\boldsymbol{e}\sim\mathcal{D}}\left[\sum_{t=1}^{L}\log\left(\boldsymbol{\pi}_{\boldsymbol{\theta}}(a_t|s_t)\right)\right]    
\end{equation}

\noindent \textbf{Chain-of-thought(CoT) Sampling}

For labeled data $D = \{q, a^-, a^+\}$, we prompt the LLM to generate multiple sampled judgments. During each generation, we use different prompts, random seeds, and temperature parameters to encourage the LLM to think and generate natural language judgments $j = \pi(p)$ from various perspectives as randomly as possible. These judgments are presented in JSON format, containing the judgment and its rationale: a key named ``CoT'' that captures the model's deep reasoning and validation process for the answers, and another key ``Chosen answer'' that indicates the LLM's final judgment. Throughout this process, the LLM is not informed which answer is superior and generates judgments in the same JSON format. In this way, we can generate multiple judgment chain-of-thought samples for any labeled data.

Simultaneously, we assign a preference score to each chain of thought, based on the perplexity (PPL) during reasoning.

\noindent \textbf{Perplexity-Based Scoring Mechanism}

Perplexity (PPL) is a widely used metric to evaluate the uncertainty of a language model when generating text \cite{jurafsky2009speech, bengio2003neural}. It originates from information theory, where it is closely related to entropy \cite{shannon1948mathematical}. Given a sequence of tokens $coti = (coti_1, coti_2, ..., coti_T)$, perplexity is defined as the exponentiation of the average negative log-likelihood \cite{brown1992estimate}:

\begin{equation}
PPL(cot_i) = \exp \left( -\frac{1}{T} \sum_{t=1}^{T} \log p(coti_t | coti_{<t}) \right),
\end{equation}
where: $T$ represents the length of the generated content. And $p(coti_t | coti_{<t})$ denotes the conditional probability of the $t$-th token given the preceding tokens in the chain-of-thought.

The intuition behind perplexity is that it measures how well a probability distribution predicts a sequence. A lower perplexity indicates that the model assigns higher confidence to the observed sequence, suggesting a more coherent and fluent output \cite{mikolov2011empirical}. Conversely, a high perplexity implies greater uncertainty and potential disfluency in the generated text.
Therefore, we introduced a multi-level scoring mechanism based on perplexity (PPL) to quantify the confidence of the thought chains generated by the self-validation strategy, thereby quantifying the quality and coherence of the thought chains. Specifically, different PPL scores correspond to different thought chain qualities, and these scores are reflected in the subsequent DPO training in the form of weights, so that the original model pays more attention to high-quality thought chains.

To effectively utilize perplexity in scoring preference levels, we transform it into a normalized score within the range $[0,1]$ \cite{radford2019language}:

\begin{equation}
PPL_{score} = \exp \left( -\frac{PPL(cot_i)}{\tau} \right),
\end{equation}
where $\tau$ is a temperature parameter controlling the normalization effect. This transformation ensures that lower perplexity values (higher confidence) result in higher preference scores, while higher perplexity values (greater uncertainty) lead to lower preference scores.

\noindent \textbf{Multi-level Preference Data Construction}

Based on the preference scores, we construct a multi-level preference dataset. Then we set a threshold \( p \), which is used to categorize the judgments into four levels:

\begin{itemize}
    \item{\textbf{Strong Accept:}} The preference score is above $p$, and the judgement is correct.
    \item{\textbf{Weak Accept:}} The preference score is below $p$, and the judgement is correct.  
    \item{\textbf{Weak Reject:}} The preference score is below $p$, and the judgment is incorrect.  
    \item{\textbf{Strong Reject:}} The preference score is above $p$, but the judgment is incorrect.
\end{itemize}

For details regarding this process, please refer to Appendix A An Example of CoT Sampling.

The choice of four categories is the optimal solution we arrived at after validation in our experiments: Our initial design goal was to capture subtle hierarchical information within preference signals while ensuring the stability of model training. Compared to binary preference modeling (e.g., distinguishing only between "Accept" and "Reject"), the four-category division (Strong Accept, Weak Accept, Weak Reject, Strong Reject) introduces a richer dimension of preference intensity. This allows the model to learn more granular reward signals, thereby enhancing its generalization ability. We experimentally verified the stability and effectiveness of this classification on several base models in the subsequent experiments.

We used the preference labels of the preference dataset itself to conduct a statistical correlation hypothesis. This hypothesis states that when a task requires complex reasoning rather than superficial matching, the correct answer usually depends on the model activating deep knowledge. Therefore, those reasoning paths may be more reasonable. Especially in this work, the number of thought chains generated is large, and under the premise of a large number of samples, this hypothesis holds.

Subsequently, we construct judgment pairs using the generated judgments with chain-of-thought and preference levels. We define the set of accepted judgments and the set of rejected judgments as \( \mathcal{M}(p)^+ \) and \( \mathcal{M}(p)^- \), respectively. The judgment pairs \( \{(a^+, a^-)\} \) are constructed through the Cartesian product of the positive judgment set \( \mathcal{M}(p)^+ \) and the negative judgment set \( \mathcal{M}(p)^- \).

For instance, if we set the sampling number to 5, generating 5 judgments where 2 are accepted and 3 are rejected, we pair them to produce \( \binom{3}{2} = 6 \) preference data pairs. This approach significantly expands the original labeled data and enhances the model's in-depth reasoning of the internal preference logic within the data.

\noindent \textbf{Multi-level Direct Preference Optimization (M-DPO)}

The objective function of traditional DPO is formulated as:

\begin{equation}
L_{\mathrm{DPO}}
=
- \,
\mathbb{E}_{q,a^+,a^-}
\bigl[
\log \sigma\bigl(r(q,a^+) - r(q,a^-)\bigr)
\bigr],
\end{equation}
where \( r(q,a) \) represents the reward model's score for the answer \( a \), and \( \sigma \) is the Sigmoid function.

This formulation only differentiates between binary preferences of positive and negative samples. The reasoning processes within sample pairs are generated by the model based on labeled data, and there can be significant differences among these processes. Some processes may involve uncertainties and noise, which can affect the convergence of the model's training process and its final performance. By introducing preference levels based on perplexity, we can effectively quantify the differences in generation quality among various chains of thought and incorporate these into the training process for the model to learn. This approach also provides more fine-grained preference relationships, such as the score differences between strong acceptance and weak acceptance.

Therefore, by optimizing the classification objective using multi-level preference data and considering the level differences as weights for sample pairs, we redefine the loss function as:

\begin{align}
L_{\text{M-DPO}} &= - \ \mathbb{E}_{q,a^+,a^-} \Biggl[ w(g^+, g^-) \cdot \nonumber \\
&\quad \log \sigma \left( r(q, a^+) - r(q, a^-) \right) \Biggr],
\end{align}
$g^+$ and $g^-$ are used to define the "preference intensity" of positive and negative samples. Their calculation is based on the perplexity scores obtained during the generation of CoT, and their values are actually hyperparameters, similar to a piecewise function. the $g^+$ values for strong accept and weak accept can be set to (2, 1), respectively, and similarly, the $g^-$ values for strong reject and weak reject can be set to (-2, -1).

The weight \( w(g^+, g^-) \) is designed as:

\begin{equation}
w(g^+,g^-)=\log(1+\exp(\alpha\cdot |g^+ - g^-|)),
\end{equation}
the weight function $w(g^+, g^-)$ dynamically adjusts weights based on level differences between answers, focusing more on pairs with larger differences. 

The improved algorithm enhances the model's ability to learn preferences by using multi-level preference data and a weighting mechanism. We demonstrated the effectiveness of M-DPO in three ways: First, its weighted log-likelihood form maintains the consistency of maximum likelihood estimation under non-negative and bounded weights, enabling it to approximate true preferences with sufficient data. Second, the gradient structure of the weighted loss ensures that multi-level weights do not disrupt the convexity or quasi-convexity of the log-likelihood, guaranteeing optimizability and stable convergence during training. Lastly, by incorporating multi-level preference distinction, the model can quickly focus on critical samples and achieve better generalization in limited and noisy data environments. For detailed mathematical derivations, please refer the appendix.

\section{Experiments}

\subsection{Baselines and Datasets}

We selected Llama-3-8B-Instruct \citep{llama3modelcard} as the base model for training the reward model using the method we introduced .

We also compared our models with other scalar reward models and generative reward models trained with different architectures. Scalar reward models including BT-model and ArmoRM \citep{ArmoRM}, generative reward models includes Auto-J 
 \citep{li2023generative}, Prometheus2 \citep{kim2024prometheus} and GenRM\citep{mahan2024generative}.  We also compared with commercial closed-source models such as GPT-4o\footnote{https://openai.com/index/hello-gpt-4o/} and Claude-3-Sonnet\citep{anthropic2024claude3}. Furthermore, we also tried to use other open source large language models(Qwen1.5-7b, Qwen2-7b \citep{qwen},Gemma-7b \citep{team2024gemma} and Mistral-7B\citep{jiang2023mistral})as the base model of our reward model for comparison, demonstrating the scalability of our framework.

In the validation on the General domain dataset, our model was trained on the publicly available Skywork-Reward-Preference-80K-v0.2 \citep{liu2024skywork} dataset and its performance was evaluated on the mainstream reward model benchmark: UltraFeedback\citep{cui2023ultrafeedback}, PKU-SafeRLHF\citep{ji2024beavertails,
ji2024pku} and Reward-Bench \citep{lambert2024rewardbench}. We selected 3,000 high-quality preference samples from the Skywork-Reward-Preference-80K-v0.2 dataset to form the training set for model training. Through data filtering, we removed any overlapping prompts between the training and test sets to ensure that the model had not encountered any test data during the training phase.

To simulate real-world low-resource domain preference modeling tasks, we constructed a medical domain preference dataset based on the iCliniq dataset\footnote{https://www.icliniq.com/} . We compile and filter a total of 3,500 entries , with 3,000 used for model training and 500 reserved for validation. To ensure the model’s ability to learn and generalize within a targeted medical domain, we primarily constrain the data content to a specific specialty or disease area. The data underwent data preprocessing steps such as deduplication and normalization, and was formatted into pair-wise preference data in the form of (question, accepted answer, rejected answer) for model training and validation.

\subsection{Inference Strategies and Hyperparameters}


We trained our model using the DeepSpeed library \citep{rasley2020deepspeed}, Zero Redundancy Optimizer (ZeRO) Stage 2 \citep{rajbhandari2020zero}, gradient checkpointing \citep{chen2016training}, HuggingFace Accelerate, and FlashAttention2 \citep{dao2023flashattention}. Our model employed mixed precision computation with bfloat16 (BF16) and tensorfloat32 (TF32). During the supervised fine-tuning (SFT) stage, we utilized the AdamW optimizer \citep{loshchilov2017fixing} with a total of 5 epochs and a warm-up ratio of 10\%. The batch size was set to 48, with a learning rate of 1e-5 and a maximum sequence length of 1024. In the model training stage, we set the peak learning rate to 1e-5 and employed a cosine scheduler for learning rate adjustment. The maximum sequence length was 4,096, and the batch size was set to 128. During the inference phase, we used VLLM \citep{kwon2023efficient} for model inference. In the chain-of-thought-driven sampling process, we adopted greedy sampling with a top-p value of 0.9, top-k value of 20, temperature of 1.0, maximum length of 512, and a repetition penalty of 1.2.

\begin{table*}[h!]
\setlength{\tabcolsep}{4pt}
    \centering
    \begin{tabular}{lccccc}
    \toprule
\multirow{2}{*}{\textbf{Category}} & \multirow{2}{*}{\textbf{Method}} & \multirow{2}{*}{\textbf{Reward-Bench}} 
          & \textbf{Ultra-} & \textbf{PKU-} & \textbf{Med-domain}\\
         & & & \textbf{Feedback} & \textbf{SafeRLHF} & \textbf{Dataset}\\
    \midrule
    \multirow{2}{*}{Scalar Model}
& BT-model & 61.2 & 46.7 & 44.2 & 68.0   \\
& ArmoRM & 62.8 & 50.1  & 52.6 & 66.0 \\
    \midrule
\multirow{4}{*}{Generative Judge}
    & Llama3-8B-Instruct & 65.0 & 62.9 & 66.4 & 51.0 \\ 
    & Auto-J & 62.3 & 63.9 & 66.9 & 60.0 \\ 
& Prometheus2 & 72.0 & 63.3 & 63.0 & 63.0 \\
& GenRM & 75.0 & 72.0 & 71.8 & 65.0\\ 
    \midrule
\multirow{2}{*}{Closed-source Model}
    & GPT-4o & 80.1 & 72.2 & 69.6 & 81.0  \\ 
& Claude-3-Sonnet & 75.2 & 71.4 & 68.3 & 80.0 \\
\midrule
\multirow{1}{*}{Ours}
& \textbf{GFRIEND} & \textbf{80.4} & \textbf{72.4} &\textbf{72.1} &\textbf{83.0} \\
    \bottomrule
    \end{tabular}
        \caption{Accuracy of models' judges on the test sets of Reward-Bench, Ultra-Feedback, PKU-SafeRLHF and Medical-domain Dataset. Scalar Model and GFRIEND are trained on 3000 samples based on Llama3-8B-Instruct. }
    \label{tab1}
\end{table*}

\begin{table*}[h!]
    \centering
    \begin{tabular}{lccccccc}
    \toprule
    \multirow{2}{*}{\textbf{Model}} & \multirow{2}{*}{\textbf{Method}} 
          & \textbf{Ultra-} & \textbf{PKU-} 
        & \multicolumn{4}{c}{\textbf{Reward-Bench}} \\
        \cmidrule(lr){5-8}
         & & \textbf{Feedback} & \textbf{SafeRLHF} 
        & \textbf{Chat} & \textbf{Chat-H} & \textbf{Safety} & \textbf{Reasoning} \\
    \midrule
    \multirow{3}{*}{Llama3-8B}
    & SFT & 62.9 & 66.4 & 85.5 & 41.6 & 68.0 & 64.8 \\ 
& BT-model & 46.7 & 44.2 & 64.3 & 58.1 & 65.2 & 57.0 \\
& \textbf{GFRIEND} & \textbf{72.4} &\textbf{72.1} & \textbf{91.9} & \textbf{62.7} & \textbf{84.5} & \textbf{82.3} \\
    \midrule
\multirow{3}{*}{Qwen1.5-7B}
    & SFT & 59.3  &  63.8 &    83.2 &  42.3&   66.5&    60.8 \\ 
& BT-model & 44.8  &  43.6  &   62.7 &  56.9 &  64.1 &   55.4 \\
& \textbf{GFRIEND} & \textbf{70.5} &\textbf{71.2} & \textbf{89.1} & \textbf{61.6} & \textbf{81.2} & \textbf{78.4} \\ 
    \midrule
\multirow{3}{*}{Qwen2-7B}
    & SFT & 60.1  &  64.5   &  84.3 &  41.8 &  67.0 &   61.2 \\ 
& BT-model & 45.6  &  44.1  &   63.9 &  57.6  & 65.4 &   56.1 \\
& \textbf{GFRIEND} & \textbf{71.3} &\textbf{72.0} & \textbf{90.2} & \textbf{62.3} & \textbf{82.4} & \textbf{79.6} \\ 
\midrule
\multirow{3}{*}{Mistral-7B}
    & SFT & 62.2  &  66.3  &   85.4  & 41.9 &  68.5  &  63.1 \\ 
& BT-model & 47.1  &  45.4  &   65.2  & 58.7  & 66.3   & 57.8 \\
& \textbf{GFRIEND} & \textbf{73.2} &\textbf{73.5} & \textbf{91.5} & \textbf{63.6} & \textbf{84.1} & \textbf{81.8} \\
    \bottomrule
    \end{tabular}
        \caption{Evaluation of different language model bases using supervised fine-tuning (SFT), BTmodel, and the GFRIEND method on the three benchmarks: UltraFeedback, PKU-SafeRLHF and Reward-Bench. With the exception of SFT, the data used to train the model were all 3000 samples.}
    \label{tab2}
\end{table*}

\subsection{Main Results}

\subsubsection{General Domain validation}

We trained our model on the publicly available dataset Skywork-Reward-Preference-80K-v0.2 and evaluated its performance on the public benchmark Reward-bench, Ultra-Feedback and PKU-SafeRLHF, as shown in Table \ref{tab1}. Our model was trained with a small dataset (approximately 3,000 samples) and outperformed other reward model frameworks trained with the same amount of data in all tests. In some cases, the benchmark performance using this small dataset was on par with or even superior to existing publicly available reward models trained on the full dataset, including commercial instruction-tuned models and LLMs specifically designed for preference judgment (such as Auto-J, Prometheus2 and GenRM). Moreover, our model achieved performance comparable to the closed-source model GPT-4o and Claude-3-Sonnet  in the benchmark tests.

To further validate the scalability of our framework, we also conducted training and validation on other large language model (LLM) bases by using 3,000 samples, with results shown in the Table \ref{tab2}. Our framework demonstrated superior efficiency and performance compared to the original reward model training frameworks.

\subsubsection{Validation of Domain-specific Datasets}

We also conducted model training and evaluation on the medical domain datasets mentioned in Section 4.1, with results shown in the Table \ref{tab1}. It can be seen that the existing frameworks have difficulty in making accurate judgments on small domain-specific datasets. In contrast, our model, which integrates chain-of-thought reasoning and judgment, achieves high accuracy even with limited data.

\subsection{Ablation Studies}

To further validate the effectiveness of our framework, we conducted ablation studies. Given that our method employs a combination of supervised fine-tuning (SFT) with chain-of-thought and multi-level direct preference optimization (M-DPO), we developed a model trained solely using the SFT loss for judgment. Additionally, we developed another variant that uses chain-of-thought sampling with standard DPO on top of SFT for comparison. As shown in Table \ref{Ablation Studies}, models trained with our framework outperformed variants that did not use multi-level DPO loss. Moreover, our model also outperformed the variant that used only SFT without chain-of-thought sampling.

\begin{table}[h!]
    \centering
    \begin{tabular}{lc}
        \hline
        Model  & Accuracy  \\
        \hline
        SFT     &  51.6\%      \\
        BT-model     &   68.5\%     \\
        GFRIEND (w/o M-DPO)   & 66.2\%      \\
        GFRIEND (w/o CoT-S) & 67.9\%     \\
        \textbf{GFRIEND (w/. M-DPO,CoT-S)} & \textbf{83.9\%}     \\
        \hline
    \end{tabular}
    \caption{Judgment accuracy of GFRIEND and its variants. CoT-S indicates whether or not to use Preference Refinement. With the exception of SFT, the models are all trained on 3000 samples based on Llama3-8B-Instruct.}
    \label{Ablation Studies}
\end{table}

Relying solely on DPO for measuring preference loss may fail to capture the deep logic within samples, especially when the sample size is small. In contrast, chain-of-thought sampling leverages the language model's inherent capability to capture deep features, enabling it to learn associations and preferences among samples even with limited data, thereby making accurate judgments.

Furthermore, using multi-level DPO to measure loss provides a more fine-grained differentiation of preference levels, guiding the model to reason and judge data in a more accurate and reasonable direction. This approach mitigates the impact of randomness and uncertainty that may arise during the model's own sampling process, which could affect the final data distribution. This perspective was also validated in our ablation studies.

In the ablation study, we compared the effects of fixed weights and no weights. Although our method performs excellently in low-resource environments, its fundamental idea does not rely on specific domain knowledge~\cite{zhao2018smart}, thus it has strong generalization ability. In addition, the method also has the potential to be extended to larger-scale, higher-quality preference datasets. Hierarchical preference modeling and weighted loss mechanisms can be extended to more levels, or even to continuous preference scores, thereby being applicable to complex dialogues and generation tasks.

\section{Discussions and Conclusions}

This study presents GFRIEND, a novel framework that addresses the challenges of reward model optimization in low-resource settings. By integrating chain-of-thought (CoT) sampling, multi-level preference modeling, and perplexity-based weighting, GFRIEND effectively enhances the performance with minimal labeled samples. The experimental results validate the framework’s capability to achieve performance comparable to models trained on extensive datasets, underscoring its potential for scalable and efficient RLHF.

A insight from this research is the importance of leveraging the inherent reasoning capabilities of LLMs~\cite{zhao2025medrag,zhao2025smart} to generate high-quality preference data. Through preference refinement, the framework captures nuanced distinctions in human preferences, while the multi-level preference modeling refines the reward model’s ability to prioritize meaningful differences in sample quality. The perplexity-based weighting further ensures that training emphasizes high-confidence samples, reducing the impact of noise and uncertainty in the data.

In the domain-specific context, GFRIEND demonstrates exceptional adaptability, achieving high accuracy even with limited samples from proprietary datasets such as medical consultations. This highlights the framework’s practicality for applications where data collection is constrained by privacy, cost, or availability. 

\newpage
\newpage

\newpage


\begin{thebibliography}{56}
\providecommand{\natexlab}[1]{#1}

\bibitem[{AI@Meta(2024)}]{llama3modelcard}
AI@Meta. 2024.
\newblock \href {https://github.com/meta-llama/llama3/blob/main/MODEL_CARD.md}
  {Llama 3 model card}.

\bibitem[{Ankner et~al.(2024)Ankner, Paul, Cui, Chang, and
  Ammanabrolu}]{ankner2024critiqueoutloud}
Zachary Ankner, Mansheej Paul, Brandon Cui, Jonathan~D Chang, and Prithviraj
  Ammanabrolu. 2024.
\newblock Critique-out-loud reward models.
\newblock \emph{arXiv preprint arXiv:2408.11791}.

\bibitem[{Anthropic(2024)}]{anthropic2024claude3}
Anthropic. 2024.
\newblock \href
  {https://www-cdn.anthropic.com/de8ba9b01c9ab7cbabf5c33b80b7bbc618857627/Model_Card_Claude_3.pdf}
  {Claude 3 model family: Opus, sonnet, haiku}.

\bibitem[{Askell et~al.(2021)Askell, Bai, Chen, Drain, Ganguli, Jones, Joseph,
  Kadavath, Kernion, Kravec et~al.}]{askell2021general}
Amanda Askell, Yuntao Bai, Anna Chen, Dawn Drain, Deep Ganguli, Andy Jones,
  Nicholas Joseph, Saurav Kadavath, Jackson Kernion, Shauna Kravec, et~al.
  2021.
\newblock A general language assistant as a laboratory for alignment.
\newblock \emph{arXiv preprint arXiv:2112.00861}.

\bibitem[{Bai et~al.(2023)Bai, Bai, Chu, Cui, Dang, Deng, Fan, Ge, Han, Huang,
  Hui, Ji, Li, Lin, Lin, Liu, Liu, Lu, Lu, Ma, Men, Ren, Ren, Tan, Tan, Tu,
  Wang, Wang, Wang, Wu, Xu, Xu, Yang, Yang, Yang, Yang, Yao, Yu, Yuan, Yuan,
  Zhang, Zhang, Zhang, Zhang, Zhou, Zhou, Zhou, and Zhu}]{qwen}
Jinze Bai, Shuai Bai, Yunfei Chu, Zeyu Cui, Kai Dang, Xiaodong Deng, Yang Fan,
  Wenbin Ge, Yu~Han, Fei Huang, Binyuan Hui, Luo Ji, Mei Li, Junyang Lin, Runji
  Lin, Dayiheng Liu, Gao Liu, Chengqiang Lu, Keming Lu, Jianxin Ma, Rui Men,
  Xingzhang Ren, Xuancheng Ren, Chuanqi Tan, Sinan Tan, Jianhong Tu, Peng Wang,
  Shijie Wang, Wei Wang, Shengguang Wu, Benfeng Xu, Jin Xu, An~Yang, Hao Yang,
  Jian Yang, Shusheng Yang, Yang Yao, Bowen Yu, Hongyi Yuan, Zheng Yuan,
  Jianwei Zhang, Xingxuan Zhang, Yichang Zhang, Zhenru Zhang, Chang Zhou,
  Jingren Zhou, Xiaohuan Zhou, and Tianhang Zhu. 2023.
\newblock Qwen technical report.
\newblock \emph{arXiv preprint arXiv:2309.16609}.

\bibitem[{Bai et~al.(2022)Bai, Jones, Ndousse, Askell, Chen, DasSarma, Drain,
  Fort, Ganguli, Henighan et~al.}]{bai2022a}
Yuntao Bai, Andy Jones, Kamal Ndousse, Amanda Askell, Anna Chen, Nova DasSarma,
  Dawn Drain, Stanislav Fort, Deep Ganguli, Tom Henighan, et~al. 2022.
\newblock Training a helpful and harmless assistant with reinforcement learning
  from human feedback.
\newblock \emph{arXiv preprint arXiv:2203.02155}.

\bibitem[{Bengio et~al.(2003)Bengio, Ducharme, Vincent, and
  Jauvin}]{bengio2003neural}
Yoshua Bengio, Réjean Ducharme, Pascal Vincent, and Christian Jauvin. 2003.
\newblock A neural probabilistic language model.
\newblock \emph{Journal of Machine Learning Research}, 3:1137--1155.

\bibitem[{Bradley and Terry(1952)}]{bradley1952rank}
Ralph~Allan Bradley and Milton~E Terry. 1952.
\newblock Rank analysis of incomplete block designs: I. the method of paired
  comparisons.
\newblock \emph{Biometrika}, 39(3/4):324--345.

\bibitem[{Brown et~al.(1992)Brown, Pietra, Mercer, Pietra, and
  Lai}]{brown1992estimate}
Peter~F. Brown, Vincent J.~Della Pietra, Robert~L. Mercer, Stephen A.~Della
  Pietra, and Jennifer~C. Lai. 1992.
\newblock An estimate of an upper bound for the entropy of english.
\newblock In \emph{Proceedings of the Workshop on Speech and Natural Language},
  pages 103--108.

\bibitem[{Chen et~al.(2016)Chen, Xu, Zhang, and Guestrin}]{chen2016training}
Tianqi Chen, Bing Xu, Chiyuan Zhang, and Carlos Guestrin. 2016.
\newblock Training deep nets with sublinear memory cost.
\newblock \emph{arXiv preprint arXiv:1604.06174}.

\bibitem[{Christiano et~al.(2017)Christiano, Leike, Brown, Martic, Legg, and
  Amodei}]{christiano2017deep}
Paul~F Christiano, Jan Leike, Tom~B Brown, Miljan Martic, Shane Legg, and Dario
  Amodei. 2017.
\newblock Deep reinforcement learning from human preferences.
\newblock \emph{Advances in Neural Information Processing Systems}, 30.

\bibitem[{Cui et~al.(2023)Cui, Yuan, Ding, Yao, Zhu, Ni, Xie, Liu, and
  Sun}]{cui2023ultrafeedback}
Ganqu Cui, Lifan Yuan, Ning Ding, Guanming Yao, Wei Zhu, Yuan Ni, Guotong Xie,
  Zhiyuan Liu, and Maosong Sun. 2023.
\newblock \href {https://arxiv.org/abs/2310.01377} {Ultrafeedback: Boosting
  language models with high-quality feedback}.
\newblock \emph{Preprint}, arXiv:2310.01377.

\bibitem[{Dao(2023)}]{dao2023flashattention}
Tri Dao. 2023.
\newblock Flashattention-2: Faster attention with better parallelism and work
  partitioning.
\newblock \emph{arXiv preprint arXiv:2307.08691}.

\bibitem[{Das et~al.(2024)Das, Chakraborty, Pacchiano, and
  Chowdhury}]{das2024active}
Nirjhar Das, Souradip Chakraborty, Aldo Pacchiano, and Sayak~Ray Chowdhury.
  2024.
\newblock Active preference optimization for sample efficient rlhf.
\newblock \emph{arXiv preprint arXiv:2402.10500}.

\bibitem[{Dong et~al.(2023{\natexlab{a}})Dong, Xiong, Goyal, Zhang, Chow, Pan,
  Diao, Zhang, Shum, and Zhang}]{dong2023raft}
Hanze Dong, Wei Xiong, Deepanshu Goyal, Yihan Zhang, Winnie Chow, Rui Pan,
  Shizhe Diao, Jipeng Zhang, Kashun Shum, and Tong Zhang. 2023{\natexlab{a}}.
\newblock Raft: Reward ranked finetuning for generative foundation model
  alignment.
\newblock \emph{arXiv preprint arXiv:2304.06767}.

\bibitem[{Dong et~al.(2024)Dong, Xiong, Pang, Wang, Zhao, Zhou, Jiang, Sahoo,
  Xiong, and Zhang}]{dong2024rlhf}
Hanze Dong, Wei Xiong, Bo~Pang, Haoxiang Wang, Han Zhao, Yingbo Zhou, Nan
  Jiang, Doyen Sahoo, Caiming Xiong, and Tong Zhang. 2024.
\newblock Rlhf workflow: From reward modeling to online rlhf.
\newblock \emph{arXiv preprint arXiv:2405.07863}.

\bibitem[{Dong et~al.(2023{\natexlab{b}})Dong, Yu, Xu, Zeng, and
  Huang}]{dong2023survey}
Li~Dong, Zhixiong Yu, Shaohan Xu, Michael Zeng, and Xuedong Huang.
  2023{\natexlab{b}}.
\newblock A survey on in-context learning.
\newblock \emph{arXiv preprint arXiv:2301.00234}.

\bibitem[{Hosseini et~al.(2024)Hosseini, Yuan, Malkin, Courville, Sordoni, and
  Agarwal}]{hosseini2024vstar}
Arian Hosseini, Xingdi Yuan, Nikolay Malkin, Aaron Courville, Alessandro
  Sordoni, and Rishabh Agarwal. 2024.
\newblock V-star: Training verifiers for self-taught reasoners.
\newblock \emph{arXiv preprint arXiv:2403.09629}.

\bibitem[{Ji et~al.(2024{\natexlab{a}})Ji, Hong, Zhang, Chen, Dai, Zheng, Qiu,
  Li, and Yang}]{ji2024pku}
Jiaming Ji, Donghai Hong, Borong Zhang, Boyuan Chen, Josef Dai, Boren Zheng,
  Tianyi Qiu, Boxun Li, and Yaodong Yang. 2024{\natexlab{a}}.
\newblock Pku-saferlhf: Towards multi-level safety alignment for llms with
  human preference.
\newblock \emph{arXiv preprint arXiv:2406.15513}.

\bibitem[{Ji et~al.(2024{\natexlab{b}})Ji, Liu, Dai, Pan, Zhang, Bian, Chen,
  Sun, Wang, and Yang}]{ji2024beavertails}
Jiaming Ji, Mickel Liu, Josef Dai, Xuehai Pan, Chi Zhang, Ce~Bian, Boyuan Chen,
  Ruiyang Sun, Yizhou Wang, and Yaodong Yang. 2024{\natexlab{b}}.
\newblock Beavertails: Towards improved safety alignment of llm via a
  human-preference dataset.
\newblock \emph{Advances in Neural Information Processing Systems}, 36.

\bibitem[{Jiang et~al.(2023)Jiang, Sablayrolles, Mensch, Bamford, Chaplot,
  Casas, Bressand, Lengyel, Lample, Saulnier et~al.}]{jiang2023mistral}
Albert~Q Jiang, Alexandre Sablayrolles, Arthur Mensch, Chris Bamford,
  Devendra~Singh Chaplot, Diego de~las Casas, Florian Bressand, Gianna Lengyel,
  Guillaume Lample, Lucile Saulnier, et~al. 2023.
\newblock Mistral 7b.
\newblock \emph{arXiv preprint arXiv:2310.06825}.

\bibitem[{Jurafsky and Martin(2009)}]{jurafsky2009speech}
Daniel Jurafsky and James~H. Martin. 2009.
\newblock \emph{Speech and Language Processing}.
\newblock Pearson.

\bibitem[{Kim et~al.(2024)Kim, Suk, Longpre, Lin, Shin, Welleck, Neubig, Lee,
  Lee, and Seo}]{kim2024prometheus}
Seungone Kim, Juyoung Suk, Shayne Longpre, Bill~Yuchen Lin, Jamin Shin, Sean
  Welleck, Graham Neubig, Moontae Lee, Kyungjae Lee, and Minjoon Seo. 2024.
\newblock Prometheus 2: An open source language model specialized in evaluating
  other language models.
\newblock \emph{arXiv preprint arXiv:2405.01535}.

\bibitem[{Kojima et~al.(2022)Kojima, Gu, Reid, Matsuo, and
  Iwasawa}]{kojima2022large}
Takeshi Kojima, Shixiang~Shane Gu, Machel Reid, Yutaka Matsuo, and Yusuke
  Iwasawa. 2022.
\newblock Large language models are zero-shot reasoners.
\newblock \emph{Advances in neural information processing systems},
  35:22199--22213.

\bibitem[{Kwon et~al.(2023)Kwon, Li, Zhuang, Sheng, Zheng, Yu, Gonzalez, Zhang,
  and Stoica}]{kwon2023efficient}
Woosuk Kwon, Zhuohan Li, Siyuan Zhuang, Ying Sheng, Lianmin Zheng, Cody~Hao Yu,
  Joseph Gonzalez, Hao Zhang, and Ion Stoica. 2023.
\newblock Efficient memory management for large language model serving with
  pagedattention.
\newblock In \emph{Proceedings of the 29th Symposium on Operating Systems
  Principles}, pages 611--626.

\bibitem[{Lambert et~al.(2024)Lambert, Pyatkin, Morrison, Miranda, Lin, Chandu,
  Dziri, Kumar, Zick, Choi, Smith, and Hajishirzi}]{lambert2024rewardbench}
Nathan Lambert, Valentina Pyatkin, Jacob Morrison, LJ~Miranda, Bill~Yuchen Lin,
  Khyathi Chandu, Nouha Dziri, Sachin Kumar, Tom Zick, Yejin Choi, Noah~A
  Smith, and Hannaneh Hajishirzi. 2024.
\newblock Rewardbench: Evaluating reward models for language modeling.
\newblock \emph{arXiv preprint arXiv:2402.05070}.

\bibitem[{Li et~al.(2023{\natexlab{a}})Li, Sun, Yuan, Fan, Zhao, and
  Liu}]{li2023generative}
Junlong Li, Shichao Sun, Weizhe Yuan, Run-Ze Fan, Hai Zhao, and Pengfei Liu.
  2023{\natexlab{a}}.
\newblock Generative judge for evaluating alignment.
\newblock \emph{arXiv preprint arXiv:2310.05470}.

\bibitem[{Li et~al.(2023{\natexlab{b}})}]{li2023instruction}
Yujia Li et~al. 2023{\natexlab{b}}.
\newblock Instruction tuning for language models.
\newblock \emph{arXiv preprint arXiv:2304.07721}.

\bibitem[{Liu et~al.(2024)Liu, Zeng, Liu, Yan, He, Wang, Yan, Liu, and
  Zhou}]{liu2024skywork}
Chris~Yuhao Liu, Liang Zeng, Jiacai Liu, Rui Yan, Jujie He, Chaojie Wang,
  Shuicheng Yan, Yang Liu, and Yahui Zhou. 2024.
\newblock Skywork-reward: Bag of tricks for reward modeling in llms.
\newblock \emph{arXiv preprint arXiv:2410.18451}.

\bibitem[{Loshchilov et~al.(2017)Loshchilov, Hutter
  et~al.}]{loshchilov2017fixing}
Ilya Loshchilov, Frank Hutter, et~al. 2017.
\newblock Fixing weight decay regularization in adam.
\newblock \emph{arXiv preprint arXiv:1711.05101}, 5.

\bibitem[{Luong et~al.(2024)Luong, Zhang, Jie, Sun, Jin, and
  Li}]{luong2024reft}
Trung~Quoc Luong, Xinbo Zhang, Zhanming Jie, Peng Sun, Xiaoran Jin, and Hang
  Li. 2024.
\newblock Reft: Reasoning with reinforced fine-tuning.
\newblock \emph{arXiv preprint arXiv:2401.08967}.

\bibitem[{Mahan et~al.(2024)Mahan, Van~Phung, Rafailov, Blagden, Lile,
  Castricato, Fr{\"a}nken, Finn, and Albalak}]{mahan2024generative}
Dakota Mahan, Duy Van~Phung, Rafael Rafailov, Chase Blagden, Nathan Lile, Louis
  Castricato, Jan-Philipp Fr{\"a}nken, Chelsea Finn, and Alon Albalak. 2024.
\newblock Generative reward models.
\newblock \emph{arXiv preprint arXiv:2410.12832}.

\bibitem[{Mikolov et~al.(2011)Mikolov, Kombrink, Burget, Černocký, and
  Khudanpur}]{mikolov2011empirical}
Tomas Mikolov, Stefan Kombrink, Lukáš Burget, Jan Černocký, and Sanjeev
  Khudanpur. 2011.
\newblock Empirical evaluation and combination of advanced language modeling
  techniques.
\newblock \emph{INTERSPEECH}.

\bibitem[{Ouyang et~al.(2022)Ouyang, Wu, Jiang, Almeida, Wainwright, Mishkin,
  Zhang, Agarwal, Slama, Ray et~al.}]{ouyang2022training}
Long Ouyang, Jeffrey Wu, Xu~Jiang, Diogo Almeida, Carroll~L. Wainwright, Pamela
  Mishkin, Chong Zhang, Saurabh Agarwal, Karel Slama, Anirudha Ray, et~al.
  2022.
\newblock Training language models to follow instructions with human feedback.
\newblock \emph{arXiv preprint arXiv:2203.02155}.

\bibitem[{Radford et~al.(2019)Radford, Wu, Child, Luan, Amodei, and
  Sutskever}]{radford2019language}
Alec Radford, Jeffrey Wu, Rewon Child, David Luan, Dario Amodei, and Ilya
  Sutskever. 2019.
\newblock Language models are unsupervised multitask learners.
\newblock \emph{OpenAI Blog}, 1:8.

\bibitem[{Rafailov et~al.(2023)Rafailov, Sharma, Mitchell, Manning, Ermon, and
  Finn}]{rafailov2023direct}
Rafael Rafailov, Archit Sharma, Eric Mitchell, Christopher~D Manning, Stefano
  Ermon, and Chelsea Finn. 2023.
\newblock Direct preference optimization: Your language model is secretly a
  reward model.

\bibitem[{Rajbhandari et~al.(2020)Rajbhandari, Rasley, Ruwase, and
  He}]{rajbhandari2020zero}
Samyam Rajbhandari, Jeff Rasley, Olatunji Ruwase, and Yuxiong He. 2020.
\newblock Zero: Memory optimizations toward training trillion parameter models.
\newblock In \emph{SC20: International Conference for High Performance
  Computing, Networking, Storage and Analysis}, pages 1--16. IEEE.

\bibitem[{Rasley et~al.(2020)Rasley, Rajbhandari, Ruwase, and
  He}]{rasley2020deepspeed}
Jeff Rasley, Samyam Rajbhandari, Olatunji Ruwase, and Yuxiong He. 2020.
\newblock Deepspeed: System optimizations enable training deep learning models
  with over 100 billion parameters.
\newblock In \emph{Proceedings of the 26th ACM SIGKDD International Conference
  on Knowledge Discovery \& Data Mining}, pages 3505--3506.

\bibitem[{Schulman et~al.(2017)Schulman, Wolski, Dhariwal, Radford, and
  Klimov}]{schulman2017proximal}
John Schulman, Filip Wolski, Prafulla Dhariwal, Alec Radford, and Oleg Klimov.
  2017.
\newblock Proximal policy optimization algorithms.
\newblock \emph{arXiv preprint arXiv:1707.06347}.

\bibitem[{Shannon(1948)}]{shannon1948mathematical}
Claude~E. Shannon. 1948.
\newblock A mathematical theory of communication.
\newblock \emph{Bell System Technical Journal}, 27(3):379--423.

\bibitem[{Team et~al.(2024)Team, Mesnard, Hardin, Dadashi, Bhupatiraju, Pathak,
  Sifre, Rivi{\`e}re, Kale, Love et~al.}]{team2024gemma}
Gemma Team, Thomas Mesnard, Cassidy Hardin, Robert Dadashi, Surya Bhupatiraju,
  Shreya Pathak, Laurent Sifre, Morgane Rivi{\`e}re, Mihir~Sanjay Kale,
  Juliette Love, et~al. 2024.
\newblock Gemma: Open models based on gemini research and technology.
\newblock \emph{arXiv preprint arXiv:2403.08295}.

\bibitem[{Wang et~al.(2024{\natexlab{a}})Wang, Xiong, Xie, Zhao, and
  Zhang}]{ArmoRM}
Haoxiang Wang, Wei Xiong, Tengyang Xie, Han Zhao, and Tong Zhang.
  2024{\natexlab{a}}.
\newblock Interpretable preferences via multi-objective reward modeling and
  mixture-of-experts.
\newblock In \emph{EMNLP}.

\bibitem[{Wang et~al.(2024{\natexlab{b}})Wang, Xu, Zhou, Xiong, and
  Joty}]{wang2024direct}
Peifeng Wang, Austin Xu, Yilun Zhou, Caiming Xiong, and Shafiq Joty.
  2024{\natexlab{b}}.
\newblock Direct judgement preference optimization.
\newblock \emph{arXiv preprint arXiv:2409.14664}.

\bibitem[{Wang et~al.(2024{\natexlab{c}})Wang, Zhang, Zhang, Hu, Li, Zhang, Li,
  Wu, Wang, and Hovy}]{wang2024reinforcement}
Shuhe Wang, Shengyu Zhang, Jie Zhang, Runyi Hu, Xiaoya Li, Tianwei Zhang, Jiwei
  Li, Fei Wu, Guoyin Wang, and Eduard Hovy. 2024{\natexlab{c}}.
\newblock Reinforcement learning enhanced llms: A survey.
\newblock \emph{arXiv preprint arXiv:2412.10400}.

\bibitem[{Wei et~al.(2022)Wei, Wang, Schuurmans, and Chi}]{wei2022chain}
Jason Wei, Kai Wang, David Schuurmans, and Edward~H Chi. 2022.
\newblock Chain of thought prompting elicits reasoning in large language
  models.
\newblock \emph{arXiv preprint arXiv:2205.11916}.

\bibitem[{Xiao et~al.(2024)Xiao, Wang, Gan, Zhao, He, Tuan, Chen, Jiang, Zhao,
  and Wu}]{xiao2024comprehensive}
Wenyi Xiao, Zechuan Wang, Leilei Gan, Shuai Zhao, Wanggui He, Luu~Anh Tuan,
  Long Chen, Hao Jiang, Zhou Zhao, and Fei Wu. 2024.
\newblock A comprehensive survey of direct preference optimization: Datasets,
  theories, variants, and applications.
\newblock \emph{arXiv preprint arXiv:2410.15595}.

\bibitem[{Xu et~al.(2023)}]{xu2023online}
Yichong Xu et~al. 2023.
\newblock Online direct preference optimization.
\newblock \emph{arXiv preprint arXiv:2307.06347}.

\bibitem[{Ye et~al.(2024)Ye, Li, Li, Ai, Zhou, Shen, Yan, and
  Liu}]{ye2024beyond}
Ziyi Ye, Xiangsheng Li, Qiuchi Li, Qingyao Ai, Yujia Zhou, Wei Shen, Dong Yan,
  and Yiqun Liu. 2024.
\newblock Beyond scalar reward model: Learning generative judge from preference
  data.
\newblock \emph{arXiv preprint arXiv:2410.03742}.

\bibitem[{Zhang et~al.(2024{\natexlab{a}})Zhang, Wang, Jin, Chen, Zhang, and
  Liu}]{zhang2024prototypical}
Jinghan Zhang, Xiting Wang, Yiqiao Jin, Changyu Chen, Xinhao Zhang, and Kunpeng
  Liu. 2024{\natexlab{a}}.
\newblock Prototypical reward network for data-efficient rlhf.
\newblock \emph{arXiv preprint arXiv:2406.06606}.

\bibitem[{Zhang et~al.(2024{\natexlab{b}})}]{zhang2024fine}
Yifan Zhang et~al. 2024{\natexlab{b}}.
\newblock Fine-tuning language models with generative feedback.
\newblock \emph{arXiv preprint arXiv:2407.06347}.

\bibitem[{Zhang et~al.(2022)Zhang, Zhang, Li, and Smola}]{zhang2022automatic}
Zhuosheng Zhang, Aston Zhang, Mu~Li, and Alex Smola. 2022.
\newblock Automatic chain of thought prompting in large language models.
\newblock \emph{arXiv preprint arXiv:2210.03493}.

\bibitem[{Zhao(2021)}]{zhao2021explainable}
Xuejiao Zhao. 2021.
\newblock Explainable q\&a system based on domain-specific knowledge graph.

\bibitem[{Zhao et~al.(2021)Zhao, Chen, Xing, and Miao}]{zhao2021brain}
Xuejiao Zhao, Huanhuan Chen, Zhenchang Xing, and Chunyan Miao. 2021.
\newblock Brain-inspired search engine assistant based on knowledge graph.
\newblock \emph{IEEE Transactions on Neural Networks and Learning Systems},
  34(8):4386--4400.

\bibitem[{Zhao et~al.(2018)Zhao, Li, Tang, Gao, Bao, and Lee}]{zhao2018smart}
Xuejiao Zhao, Hongwei Li, Yutian Tang, Dongjing Gao, Lingfeng Bao, and
  Ching-Hung Lee. 2018.
\newblock A smart context-aware program assistant based on dynamic programming
  event modeling.
\newblock In \emph{ISSRE Workshops}, pages 24--29.

\bibitem[{Zhao et~al.(2025{\natexlab{a}})Zhao, Liu, Yang, and
  Miao}]{zhao2025medrag}
Xuejiao Zhao, Siyan Liu, Su-Yin Yang, and Chunyan Miao. 2025{\natexlab{a}}.
\newblock Medrag: Enhancing retrieval-augmented generation with knowledge
  graph-elicited reasoning for healthcare copilot.
\newblock In \emph{Proceedings of the ACM on Web Conference 2025}, pages
  4442--4457.

\bibitem[{Zhao et~al.(2025{\natexlab{b}})Zhao, Liu, Yang, and
  Miao}]{zhao2025smart}
Xuejiao Zhao, Siyan Liu, Su-Yin Yang, and Chunyan Miao. 2025{\natexlab{b}}.
\newblock A smart multimodal healthcare copilot with powerful llm reasoning.
\newblock \emph{arXiv preprint arXiv:2506.02470}.

\end{thebibliography}

\newpage

\appendix

\addcontentsline{toc}{section}{Appendix}
\section*{Appendix}

\section{An Example of CoT Sampling}

\begin{table*}[htbp]
  \centering
  \footnotesize              
  \caption{An example of CoT Sampling.}
  \label{tab:cot-sampling-example1}
  \begin{tabular}{@{}p{0.18\textwidth}p{0.78\textwidth}@{}}
    \toprule
    \textbf{Field} & \textbf{Content} \\
    \midrule
    Preference Data &
      \begin{tabular}[t]{@{}l@{}}
        \textbf{Question}: “On the theme of comprehensive rural revitalization, propose a paper title.”\\
        \textbf{Reject}: “Research on Strategic Pathways for Rural Revitalization from a Multi-Collaborative Perspective.”\\
        \textbf{Accept}: “Comprehensive Rural Revitalization: Deep Integration of Strategies and Practices.”
      \end{tabular} \\
    \addlinespace
    Prompt &
      “Which answer do you think is better? Please think step by step and give your reasons.” \\
    \addlinespace
    Model’s CoT Output &
      \begin{tabular}[t]{@{}l@{}}
        \textbf{CoT}: [The question emphasizes “comprehensive rural revitalization.” \\
        Answer 2 begins with “Comprehensive \\Rural Revitalization,” which precisely echoes the key feature of “comprehensive,” \\
        while Answer 1 only mentions “Rural Revitalization” without highlighting the “comprehensive” aspect, making \\ its
        relevance slightly weaker. Additionally, Answer 2 emphasizes the “deep integration of strategies and practices,” \\
        directly addressing the critical dimension for the implementation of rural revitalization and highlighting the \\practical orientation of the research.\\ 
        In contrast, Answer 1’s “Multi-Collaborative Perspective” is more methodological, focusing on collaboration \\among
        stakeholders, with a narrower thematic coverage and failure to highlight the core research dimension. 
        \\
        Furthermore, the keywords “comprehensive + deep integration” in Answer 2 make the scope and focus of the \\research clear and more appealing, while Answer 1’s “Strategic Pathways” is relatively vague,\\
        requiring readers to infer the research focus further, lacking clarity. \\
        Considering these public factors, Answer 2 has a stronger direct correspondence with the specified theme and \\clearer research focus, making it the better choice for the title.]\\
        \textbf{Chosen Answer}: [Answer 2]
      \end{tabular} \\
    \addlinespace
    PPL Score      & 0.8 \\
    Classification & Strong Accept \\
    \bottomrule
  \end{tabular}
\end{table*}

We construct an example of chain-of-thought sampling(Table \ref{tab:cot-sampling-example1}). During the forward output of this answer, the PPL score was 0.8 (the specific calculation process can be found in formulas (7) and (8) in the paper and the code repository). Since the model's judgment is consistent with the label and its score exceeds the threshold (the threshold here is 0.5), this sample is classified as a "strong accept" sample.

We also provided the output of the model that had not undergone Chain-of-Thought Supervised Fine-Tuning (CoT SFT) based on LLaMA-3-8B-Instruct when given the same preference questions(Table \ref{tab:cot-sampling-example2}). It can be observed that the output content of this model is far less rich and in-depth compared to that of the model that had undergone CoT SFT. Such content fails to reflect the implicit preference information contained in the existing labeled data, which undoubtedly highlights the importance of the CoT mechanism we designed in finely mining preference information. Moreover, it can also be seen that without specific SFT, the model is even unable to make effective preference judgments in a single inference.

\begin{table*}[h]
  \centering
  \footnotesize              
  \caption{An example of the model's output without undergoing CoT SFT.}
  \label{tab:cot-sampling-example2}
  \begin{tabular}{@{}p{0.18\textwidth}p{0.78\textwidth}@{}}
    \toprule
    \textbf{Field} & \textbf{Content} \\
    \midrule
    Preference Data &
      \begin{tabular}[t]{@{}l@{}}
        \textbf{Question}: “On the theme of comprehensive rural revitalization, propose a paper title.”\\
        \textbf{Reject}: “Research on Strategic Pathways for Rural Revitalization from a Multi-Collaborative Perspective.”\\
        \textbf{Accept}: “Comprehensive Rural Revitalization: Deep Integration of Strategies and Practices.”
      \end{tabular} \\
    \addlinespace
    Prompt &
      “Which answer do you think is better? Please think step by step and give your reasons.” \\
    \addlinespace
    Model’s CoT Output &
      \begin{tabular}[t]{@{}l@{}}
        [Given the question's context, the choice between the two proposed titles depends on the specific \\
        focus of your paper:Answer 1 is better if your paper aims to explore and\\
        analyze various strategic pathways for rural revitalization, emphasizing a collaborative approach. \\

Answer 2 is more suitable if your paper intends to discuss \\
the deep integration of strategies and practices in rural revitalization.\\ 

In summary, the choice should align with the depth of analysis, the focus on collaboration or integration, \\
and the overall approach you wish to take in your paper.]\\
      \end{tabular} \\
    \addlinespace
    PPL Score      & 0.72 \\
    \bottomrule
  \end{tabular}
\end{table*}

\section{Line Chart Analysis}

\begin{figure}[h]
\centering
\includegraphics[width=0.5\textwidth]{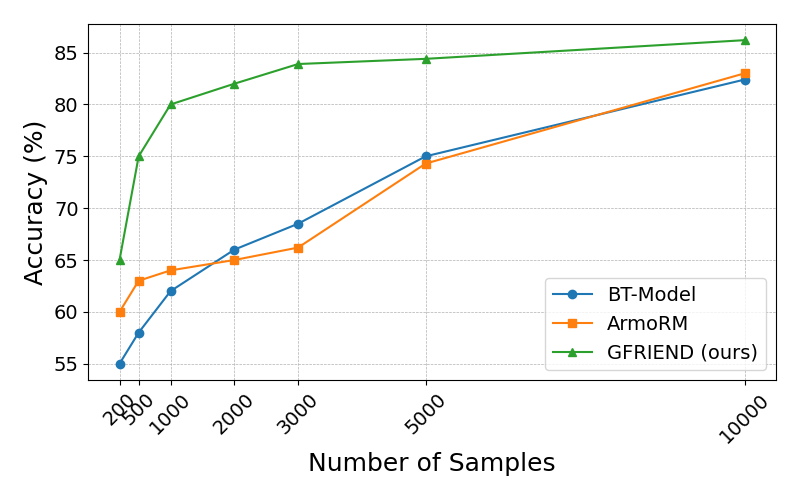}
\caption{This line chart compares the performance of different reward modeling frameworks under varying increments of data.} 
\label{fig00}
\end{figure}

We created a line chart (Figure \ref{fig00}) to evaluate the performance of different reward modeling frameworks under varying training data volumes. It can be observed that in the early stages of training, our framework's evaluation accuracy is significantly higher than that of existing reward modeling frameworks, highlighting the superior capability of our method in few-shot preference modeling tasks. The dataset used in this process is based on Skywork-Reward-Preference-80K-v0.2.

We created a line chart to evaluate the performance of different reward modeling frameworks under varying training data volumes. It can be observed that in the early stages of training, our framework's evaluation accuracy is significantly higher than that of existing reward modeling frameworks, highlighting the superior capability of our method in few-shot preference modeling tasks. The dataset used in this process is based on Skywork.

\section{Proof of M-DPO Loss Function Validity}
\noindent The following content provides a mathematical proof regarding the effectiveness of the \textbf{M-DPO (Multi-level Direct Preference Optimization)} loss function proposed in the paper, approached from the perspective of mathematical reasoning. The proof is organized into three parts, integrating insights from statistical learning theory and preference learning, to illustrate how M-DPO can simultaneously maintain consistency in finite-data scenarios (i.e., consistency in a statistical sense) and effectively capture multi-level preferences.

\textbf{Preference Data and Basic Notation}\newline Given a query $q$ and two candidate responses $a^+$ and $a^-$, we possess either human annotations or approximate annotations (e.g., generative preference judgments) indicating that $a^+$ is preferred over $a^-$. We write such a sample as $(q, a^+, a^- )$.\newline Let $r(q,a)$ denote the \textbf{reward model} score assigned to the response $a$. The traditional Bradley-Terry model assumes that\begin{equation}\label{eq:bt}\displaystyle P(a^+ \succ a^- \mid q) = \sigma\bigl(r(q,a^+) - r(q,a^- )\bigr),\end{equation}where $\sigma(x) = 1/(1 + e^{-x})$ is the sigmoid function.\newline\newline In the classical Direct Preference Optimization (DPO) framework, the objective function (ignoring any regularization terms) is given by\begin{equation}\label{eq:dpo}\displaystyle L_{\mathrm{DPO}} = \mathbb{E}_{(q,a^+,a^-)\in D}\Bigl[\log \sigma\bigl(r(q,a^+) - r(q,a^- )\bigr)\Bigr].\end{equation}\noindent Maximizing this expectation is equivalent to minimizing its negative, so we have\begin{equation}\min_{\theta} -\,\mathbb{E}_{(q,a^+,a^-)\in D}\Bigl[\log \sigma\bigl(r(q,a^+) - r(q,a^-)\bigr)\Bigr].\end{equation}

\textbf{M-DPO: Incorporating Multi-Level Preferences and Weighting}
\newline The paper introduces an extension to DPO that accounts for multi-level preferences, referred to as \textbf{M-DPO}. Let\begin{itemize}\item $g^+$ denote the "preference level" or "confidence strength" of the positive sample,\item $g^-$ denote the "preference level" or "confidence strength" of the negative sample,\end{itemize}\noindent and define\begin{equation}\label{eq:weight}\displaystyle w(g^+, g^-) = \log\bigl(1 + \exp\bigl(\alpha\cdot |g^+ - g^-|\bigr)\bigr),\end{equation}\noindent where $\alpha > 0$ is a hyperparameter controlling the sensitivity of the weight.\newline Consequently, the M-DPO objective can be written as
\begin{align}
L_{\text{M-DPO}} &= \mathbb{E}_{q,a^+,a^-} \Biggl[ w(g^+, g^-) \cdot \nonumber \\
&\quad \log \sigma \left( r(q, a^+) - r(q, a^-) \right) \Biggr]
\end{align}
\noindent Similarly, we typically minimize its negative form:\begin{equation}\displaystyle \min_{\theta}\; -\!\sum_{(q,a^+,a^- )\in D} w\bigl(g^+, g^-\bigr)\;\log \sigma\!\bigl(r(q,a^+) - r(q,a^- )\bigr).\end{equation}\noindent In what follows, we provide theoretical justifications for this objective based on consistency from statistical learning theory, weighted convexity analyses, and capturing multi-level preferences.


\textbf{Consistency Perspective}
\newline In preference learning,  consistency typically means that if the training set grows arbitrarily large and is drawn from a distribution compatible with some "true" scoring function, then the learned solution (with parameters $\theta^*$) should reproduce the actual preferences either perfectly or asymptotically perfectly.\newline For M-DPO, assume there exists a true latent scoring function $r^*(q,a)$, such that\begin{equation}\displaystyle P(a^+ \succ a^- \mid q) = \sigma\bigl(r^*(q,a^+) - r^*(q,a^-)\bigr).\end{equation}\noindent In the limit of infinite data, maximizing\begin{equation}\max_{\theta}\;\mathbb{E}\Bigl[w(g^+, g^-)\,\log \sigma\bigl(r_{\theta}(q,a^+) - r_{\theta}(q,a^-)\bigr)\Bigr]\end{equation}\noindent should yield $\theta^*$ consistent with $r^*$, i.e., $\hat{r}_{\theta^*} = c\,r^*$ (up to an additive or multiplicative constant).\newline\emph{Proof Idea:}\begin{enumerate}\item \emph{Exponential-Family Representation}\newline The Bradley-Terry model can be viewed as a logistic model within the exponential-family framework. As long as the loss function remains a weighted version of the logistic (i.e., negative log-likelihood) form, and the weighting function $w(g^+, g^-)$ is nonnegative and bounded, the statistical properties of maximum likelihood estimation (MLE) hold. Hence, if data is sufficiently large and follows the underlying true distribution, the maximum-likelihood solution is consistent.\item \emph{Nonnegativity and Boundedness of $w(g^+, g^-)$}\newline By definition,\[w(g^+, g^-) = \log\bigl(1 + \exp\bigl(\alpha \cdot |g^+ - g^-|\bigr)\bigr).\]\noindent For any real $|g^+ - g^-| \ge 0$ and $\alpha > 0$, we have\[w(g^+, g^-) > 0,\quad \text{and}\quad w(g^+, g^-) \le \log\bigl(1 + e^{\alpha\,\Delta}\bigr),\]\noindent where $\Delta$ can be considered the maximum possible gap in $|g^+ - g^-|$ (for instance, if $g^+$ and $g^-$ lie in a bounded range).\newline Thus, $w(\cdot)$ is nonnegative and bounded, which is consistent with general theorems on weighted logistic log-likelihood. Consequently, if $N\to\infty$, the weighted log-likelihood estimator remains consistent under standard i.i.d. assumptions.\end{enumerate}\noindent Thus, M-DPO does not undermine the convergence or consistency of logistic-based estimators. As the dataset size increases, the learned $\hat{r}(q,a)$ converges to the true scoring function $r^*(q,a)$ (modulo constants).
\newline\newline
\textbf{Weighted Convexity and Gradient Analysis}\newline Consider the negative log-likelihood form of M-DPO:
\begin{align}
\ell(\theta) = -\!\!\sum_{(q,a^+,a^- )\in D} w\bigl(g^+, g^-\bigr) \cdot \nonumber \\
 \log \sigma\!\bigl(r_{\theta}(q,a^+) - r_{\theta}(q,a^-)\bigr).
\end{align}

\noindent Suppose $r_{\theta}$ is a linear function of $\theta$ (e.g., the last layer is a linear output), or more generally that the neural network structure keeps $\sigma(\cdot)$ differentiable. Then the gradient of $\ell(\theta)$ follows that of the standard logistic regression, except for an extra factor $w(\cdot)$.\newline\begin{itemize}\item \textbf{(1) Standard Log-Likelihood Convexity}\newline In classical logistic regression, $-\log \sigma(y\, w^\top x)$ (where $y=\pm 1$) is a convex function in the parameters. While $\log\sigma(\Delta)$ is not necessarily globally convex in $\Delta$, it commonly possesses suitable quasi-convex properties under typical feature mappings, making it tractable.\item \textbf{(2) The Impact of $w(g^+, g^-)$}\newline In M-DPO, each sample's loss is multiplied by $w(g^+, g^-)\ge 0$. Mathematically, this corresponds to an \emph{importance weighting} mechanism on the gradient update. Provided $w(\cdot)$ does not vary wildly with changes in $\theta$ (i.e., it depends on the sample's multi-level preference labels, not on the model parameters), the usual logistic convexity or quasi-convexity remains valid.\end{itemize}\noindent As a result, the gradient update in M-DPO is essentially a multiplicative correction to the original DPO. This maintains feasibility and stability for standard optimizers like SGD or AdamW, preserving their convergence properties.
\newline\newline\textbf{Capturing Multi-Level Preferences}\newline Traditional binary preference models (e.g., $\sigma(r(q,a^+) - r(q,a^-))$) only encode which response is better, ignoring the confidence levels in $a^+$ or $a^-$. In real scenarios, the distinction between a \textbf{strongly accepted (Strong Accept)} or \textbf{weakly accepted (Weak Accept)} response can be crucial.\newline M-DPO encodes multi-level preference signals via the weighting function $w(g^+, g^-)$. Suppose\[g^+ \in \{\text{Weak Accept}, \text{Strong Accept}\}\]
\[g^- \in \{\text{Weak Reject}, \text{Strong Reject}\}.\]When $|g^+ - g^-|$ is large, it indicates a more significant preference gap between the positive and negative samples, thus the algorithm places higher weight on these samples to more rapidly widen the margin $r(q,a^+) - r(q,a^-)$. Conversely, if $|g^+ - g^-|$ is moderate (e.g., Weak Accept vs. Weak Reject), the weight is smaller. From\[w(g^+, g^-) = \log\bigl(1 + \exp\bigl(\alpha\cdot|g^+ - g^-|\bigr)\bigr),\]we see that as $|g^+ - g^-|$ grows, $w$ increases monotonically. The rate of increase is modulated by $\alpha$, aligning with our intuition to prioritize samples where the preference difference is more pronounced.\newline From an information-theoretic viewpoint, if a sample provides a high degree of certainty about correctness (a "strong signal"), its contribution to the total information gain is also high. In the weighted framework, the model prioritizes satisfying these high-certainty constraints first, enabling $r_{\theta}$ to quickly and stably learn to distinguish important preference differences.

\textbf{Conclusions}
\newline Integrating the above three angles, we can draw the following conclusions from a purely mathematical and statistical learning standpoint:\begin{enumerate}\item \textbf{Consistency}\newline M-DPO, as a weighted version of the logistic (Bradley-Terry) preference model, preserves the consistency of the underlying maximum-likelihood estimator for large sample sizes. The model thus converges to an optimal solution consistent with the true preference ordering (up to an additive scaling).\item \textbf{Optimizability and Convergence}\newline Because $w(g^+, g^-)$ is nonnegative, bounded, and not directly coupled to the model parameters, M-DPO only applies a multiplicative correction at each sample in the gradient updates. This retains the positive gradient properties of logistic log-likelihood, allowing standard optimizers to converge efficiently.\item \textbf{Effective Multi-Level Preference Representation}\newline M-DPO's weighting function incorporates multi-level preference data, capturing distinctions such as Strong Accept vs. Weak Accept or Strong Reject vs. Weak Reject. It intensifies the focus on samples with greater preference gaps, yielding more rapid and stable discrimination in high-confidence, high-contrast cases.\end{enumerate}
\noindent Consequently, \textbf{M-DPO not only inherits the theoretical consistency and convergence strengths of the original DPO in binary preference learning but further leverages adaptive weighting for multi-level preference gaps, achieving better robustness and data efficiency in scenarios with limited or noisy data.} This directly underpins the advantage observed in few-shot or low-resource settings discussed in the paper.

\end{document}